\documentclass[10pt,twocolumn,letterpaper]{article} 

\usepackage{avss}
\usepackage{times}
\usepackage{epsfig}
\usepackage{graphicx}
\usepackage{amsmath}
\usepackage{amssymb}

\usepackage{bm}
\usepackage{microtype}

\usepackage[pagebackref=true,breaklinks=true,letterpaper=true,colorlinks,bookmarks=false]{hyperref}

\avssfinalcopy 

\def\avssPaperID{0045} 

\ifavssfinal\fi
\begin{document}

\title{MultAV: Multiplicative Adversarial Videos}

\author{Shao-Yuan Lo \hspace{0.3cm} Vishal M. Patel \\
Dept. of Electrical and Computer Engineering, Johns Hopkins University\\
3400 N. Charles St, Baltimore, MD 21218, USA\\
{\tt\small \{sylo, vpatel36\}@jhu.edu}
}

\maketitle
\thispagestyle{empty}

\begin{abstract}
The majority of adversarial machine learning research focuses on additive attacks, which add adversarial perturbation to input data. On the other hand, unlike image recognition problems, only a handful of attack approaches have been explored in the video domain. In this paper, we propose a novel attack method against video recognition models, Multiplicative Adversarial Videos (MultAV), which imposes perturbation on video data by multiplication. MultAV has different noise distributions to the additive counterparts and thus challenges the defense methods tailored to resisting additive adversarial attacks. Moreover, it can be generalized to not only $\ell_p$-norm attacks with a new adversary constraint called ratio bound, but also different types of physically realizable attacks. Experimental results show that the model adversarially trained against additive attack is less robust to MultAV.
\let\thefootnote\relax\footnote{Copyright: 978-1-6654-3396-9/21/\$31.00 ©2021 IEEE}
\end{abstract}

\section{Introduction}

\begin{figure*}[htp!]
	\begin{center}
		\centering
		\includegraphics[width=1\textwidth]{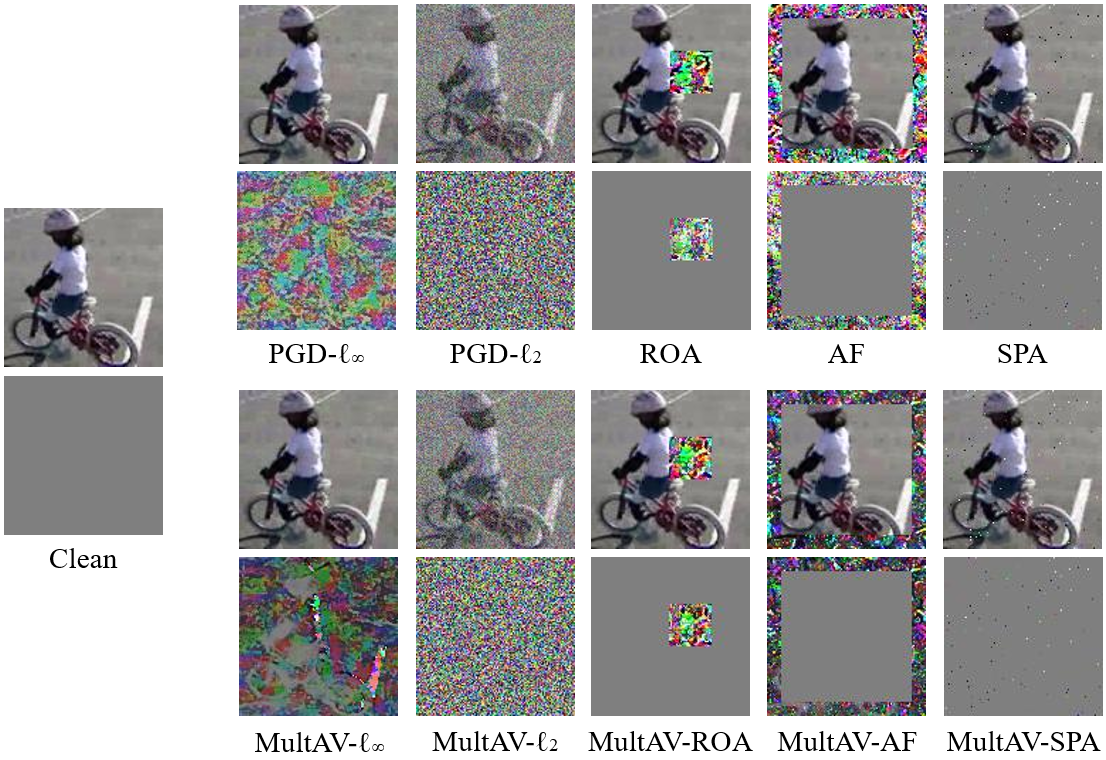}
		\caption{MultAV examples generated against 3D ResNet-18 (top); and difference maps ($15 \times$ magnified for PGD-$\ell_\infty$, PGD-$\ell_2$, MultAV-$\ell_\infty$ and MultAV-$\ell_2$) between clean and each MultAV example (bottom). The first frame of the video is displayed here. Detailed attack settings are presented in Sec. \ref{setup}.}
		\label{fig:noise}
	\end{center}
\end{figure*}

Recent advances in machine learning have led deep neural networks (DNNs) to perform extremely well in a variety of problems, but researchers have shown that DNNs are easily misled when presented by adversarial examples, causing serious security concerns. The majority of existing adversarial attacks, including $\ell_p$-bounded attacks \cite{Szegedy2014Intriguing,goodfellow2015explaining,madry2018towards} and physical attacks \cite{Sharif16AdvML,Brown2017adversarial,Wu2020Defending} emphasize the \textit{additive} attack approach where perturbation is \textit{added} to input data; that is,
\begin{equation}
\mathbf{x}_{adv} = \mathbf{x} + \bm{\delta}_{adv}.
\end{equation}
Lately, researchers have been studying some other attack types. Yang and Ji \cite{yang2019learning} presented a type of perturbation that multiplies input images by trained binary masks, but such perturbation is perceptible and requires complicated gradient estimators to optimize. Besides, their purpose is to regularize semi-supervised learning, so the adversarial attack strength is not evaluated. Some other recent works consider spatial attacks, which fool DNNs by small spatial perturbations \cite{xiao2018spatially,engstrom2019exploring}; and coloring-based attacks, which generate adversarial examples by re-coloring the input images \cite{laidlaw2019functional}. Solely delving into a handful of attack approaches would make the research community overlook many other possible adversarial examples that are threatening our machine learning systems.  Hence, novel attacks need to be explored. 

On the other hand, adversarial examples for videos have not been investigated much in the literature. Wei et al. \cite{wei2019sparse} explored temporally sparse attacks against video recognition models. Zajac et al. \cite{zajac2019adversarial} attached adversarial framings on the border of video frames. Li et al. \cite{li2019stealthy} produced adversarial videos by a generative model. Jiang et al. \cite{jiang2019black} developed V-BAD against video recognition networks in the black-box setting. All of these attacks are additive attacks.

In this paper, we propose a novel attack method, Multiplicative Adversarial Videos (MultAV), which can be applied to both ratio-bounded attacks and physically realizable attacks. Many coherent imaging systems such as synthetic aperture radar (SAR) and ultrasound often suffer from multiplicative noises, known as speckle \cite{speckle}. Inspired by this noise type, MultAV generates adversarial videos by \textit{multiplying} crafted noise with input examples:
\begin{equation}
\mathbf{x}_{adv} = \mathbf{x} \odot \bm{\delta}_{adv},
\end{equation}
where $\odot$ denotes element-wise multiplication. MultAV can be imposed by different regularizations to keep the changes imperceptible using a new constraint on adversaries called ratio bound (RB). The ratio bound restricts the pixel-wise ratio of an adversarial example to an input example, corresponding to the $\ell_p$-norm of additive counterparts that restricts the pixel-wise difference. Furthermore, MultAV also applies to salt-and-pepper attack (SPA) \cite{lo2020defending} and physically realizable attacks, where we consider the video version of rectangular occlusion attack (ROA) \cite{Wu2020Defending} and adversarial framing (AF) \cite{zajac2019adversarial} in this work. We demonstrate that these attack types can be generated by the proposed multiplicative algorithm as well.

MultAV produces different perturbation distributions to the additive counterparts and thus challenges the defense approaches which are tailored to defending against additive adversarial attacks. Specifically, given MultAV examples of an attack type (the multiplicative version of this attack), the model adversarially trained against the additive counterpart is less robust than the model adversarially trained against MultAV directly. This gap also appears on Feature Denoising \cite{xie2019feature}, a state-of-the-art defense, which demonstrates the threat of our MultAV and encourages more general and robust methods.

\section{Multiplicative Adversarial Video}
We propose MultAV to fool video recognition systems. Recall that FGSM \cite{goodfellow2015explaining} builds the foundation for additive adversarial attacks, then PGD \cite{madry2018towards} extends FGSM to iterative versions for producing stronger attacks. Given a video data sample $\mathbf{x} \in \mathbb{R}^{F \times C \times H \times W}$ ($F$ is the number of video frames, $C$ is the number of channels, $H$ and $W$ are height and width), ground-truth label $\mathbf{y}$, target model parameters $\theta$ and loss function $\mathcal{L}$, these iterative FGSM-based attacks generate adversarial examples $\mathbf{x}_{adv}$ by
\begin{equation}
\label{pgd_linf}
\mathbf{x}^{t+1} = Clip^{\ell_\infty}_{\mathbf{x},\epsilon} \big\{ \mathbf{x}^t+\alpha \cdot sign(\bigtriangledown_{\mathbf{x}^t} \mathcal{L}(\mathbf{x}^t,\mathbf{y};\:\bm{\theta})) \big\},
\end{equation}
where $\alpha$ is step size, $t \in [0,T-1]$ denotes the number of attacking iterations and thus $\mathbf{x} = \mathbf{x}^0$ and $\mathbf{x}_{adv} = \mathbf{x}^T$. $Clip^{\ell_\infty}_{\mathbf{x},\epsilon}\{\cdot\}$ denotes element-wise clipping with perturbation size $\epsilon$ such that $|\mathbf{x}^{t+1}-\mathbf{x}| \leq \epsilon$. This $\ell_\infty$-norm is the initial constraint used by the FGSM-based attacks. These attacks can also be bounded in $\ell_2$-norm:
\begin{equation}
\mathbf{x}^{t+1} = Clip^{\ell_2}_{\mathbf{x},\epsilon} \big\{ \mathbf{x}^t+\alpha \cdot \frac{\bigtriangledown_{\mathbf{x}^t} \mathcal{L}(\mathbf{x}^t,\mathbf{y};\:\bm{\theta})}{\|\bigtriangledown_{\mathbf{x}^t} \mathcal{L}(\mathbf{x}^t,\mathbf{y};\:\bm{\theta})\|_2} \big\},
\end{equation}
where $Clip^{\ell_2}_{\mathbf{x},\epsilon}\{\cdot\}$ is a $\ell_2$-norm constraint with $\epsilon$ such that $\|\mathbf{x}^{t+1}-\mathbf{x}\|_2 \leq \epsilon$. In this case, the attacks take steps in the normalized gradient values instead of the sign of them.

MultAV belongs to gradient methods as well, and it can be formulated in a single step or iterative version. The iterative MultAV is defined as
\begin{equation}
\label{mult_ratio}
\mathbf{x}^{t+1} = Clip^{RB-\ell_\infty}_{\mathbf{x},\epsilon_m} \big\{ \mathbf{x}^t \odot \alpha_m^{sign(\bigtriangledown_{\mathbf{x}^t} \mathcal{L}(\mathbf{x}^t,\mathbf{y};\,\bm{\theta}))} \big\},
\end{equation}
where $\alpha_m$ is the multiplicative step size, $Clip^{RB-\ell_\infty}_{\mathbf{x},\epsilon_m}\{\cdot\}$ performs clipping with ratio bound $\epsilon_m$ such that $\max(\frac{\mathbf{x}^{t+1}}{\mathbf{x}},\frac{\mathbf{x}}{\mathbf{x}^{t+1}})  \leq \epsilon_m$. MultAV takes the sign of the gradients as the exponent of $\alpha_m$, so that $\mathbf{x}^t$ would be multiplied by either $\alpha_m$ or $1/\alpha_m$, which is corresponded to the additive counterparts added either $\alpha$ or $-\alpha$. The ratio bound is favorable to the multiplicative cases because the $\ell_\infty$-norm would biased clip more perturbation in brighter pixels (having larger pixel values). Both the ratio bound and the $\ell_\infty$-norm limit the perturbation maximum but in terms of addition and multiplication, respectively. MultAV can also be extended to an $\ell_2$-norm variant:
\begin{equation}
\label{mult_l2}
\mathbf{x}^{t+1} = Clip^{RB-\ell_2}_{\mathbf{x},\epsilon_m} \big\{ \mathbf{x}^t \odot \alpha_m^{ \frac{\bigtriangledown_{\mathbf{x}^t} \mathcal{L}(\mathbf{x}^t,\mathbf{y};\:\bm{\theta})}{\|\bigtriangledown_{\mathbf{x}^t} \mathcal{L}(\mathbf{x}^t,\mathbf{y};\:\bm{\theta})\|_2}} \big\},
\end{equation}
where $Clip^{RB-\ell_2}_{\mathbf{x},\epsilon_m}\{\cdot\}$ a $\ell_2$-norm ratio constraint with $\epsilon_m$ such that $\|\frac{\mathbf{x}^{t+1}}{\mathbf{x}}\|_2 \leq (\epsilon_m+1)$. Adding 1 is just an offest, so that we can find a proper $\epsilon_m$ value easier. In this case, MultAV takes the normalized gradient values as the exponent of $\alpha_m$.

\renewcommand{\arraystretch}{1.5}
\setlength{\tabcolsep}{8pt}
\begin{table*}[htp!]
	\begin{center}
		\caption{Video recognition accuracy (\%) on the UCF101 dataset. The \textit{Clean} column corresponds to the models that are trained and tested on clean data. The \textit{Training} column refers to the data type for training/adversarial training, where \textit{Mult} is a MultAV type corresponding to each of the last five colums, and \textit{Add} is an additive counterpart of each of the last five colums. The numbers in parentheses are the accuracy differences between Mult Model and Add Model.}
		\label{table:results}
		\begin{tabular}{l | r | c | rrrrr}
			\hline \noalign{\smallskip} \noalign{\smallskip}
			Network & Clean & Training & MultAV-$\ell_\infty$ & MultAV-$\ell_2$ & MultAV-ROA & MultAV-AF & MultAV-SPA \\
			\noalign{\smallskip} \hline \noalign{\smallskip}
			3D ResNet-18 & 76.90 & Clean & 7.19 & 2.67 & 2.30 & 0.26 & 4.02 \\
			\noalign{\smallskip} \hline \noalign{\smallskip}
			3D ResNet-18 & 76.90 & Mult & 47.00 & 16.23 & 44.12 & 66.35 & 55.54 \\
			& & Add & 41.61 & 9.94 & 42.45 & 51.23 & 54.74 \\
			& & & (\textbf{-5.39}) & (\textbf{-6.29}) & (\textbf{-1.67}) & (\textbf{-15.12}) & (\textbf{-0.80}) \\
			\noalign{\smallskip} \hline \noalign{\smallskip}
			3D ResNet-18 & 70.82 & Mult & 42.69 & 14.75 & 39.31 & 60.53 & 48.37 \\
			+ 3D Denoise &  & Add & 31.46 & 9.15 & 37.72 & 48.98 & 48.06 \\
			& & & (\textbf{-11.23}) & (\textbf{-5.60}) & (\textbf{-1.59}) & (\textbf{-11.55}) & (\textbf{-0.31}) \\			
			\noalign{\smallskip} \hline \noalign{\smallskip}
			3D ResNet-18 & 69.47 & Mult & 41.87 & 14.04 & 40.34 & 58.97 & 47.48 \\
			+ 2D Denoise &  & Add & 30.16 & 10.23 & 39.65 & 47.82 & 47.18 \\
			& & & (\textbf{-11.71}) & (\textbf{-3.81}) & (\textbf{-0.69}) & (\textbf{-11.15}) & (\textbf{-0.30}) \\	
			\noalign{\smallskip} \hline
		\end{tabular}
	\end{center}
\end{table*}

SPA \cite{lo2020defending} and the physically realizable ROA \cite{Wu2020Defending} and AF \cite{zajac2019adversarial} examples can be generated by Eq. \ref{pgd_linf}, but the perturbation is restricted in pre-defined regions, such as a rectangular, a framing and selected pixels. These attacks allow a large perturbation size since they can be perceptible. Similarily, their multiplicative versions can be produced by Eq. \ref{mult_ratio}. MultAV is a general algorithm that applies to various attack types.

The perturbation generated by MultAV has distinct properties from that of additive adversarial examples. In particular, we can rewrite Eq. \eqref{mult_ratio} as
\begin{equation}
\label{mult_ratio_add}
\mathbf{x}^{t+1} = Clip^{RB-\ell_\infty}_{\mathbf{x},\epsilon_m} \big\{ \mathbf{x}^t + \Big[ \mathbf{x}^t \odot (\alpha_m^{sign(\bigtriangledown_{\mathbf{x}^t} \mathcal{L}(\mathbf{x}^t,\mathbf{y};\,\bm{\theta}))} - 1) \Big] \big\},
\end{equation}
and rewrite Eq. \eqref{mult_l2} as
\begin{equation}
\label{mult_l2_add}
\mathbf{x}^{t+1} = Clip^{RB-\ell_2}_{\mathbf{x},\epsilon_m} \big\{ \mathbf{x}^t + \Big[ \mathbf{x}^t \odot (\alpha_m^{ \frac{\bigtriangledown_{\mathbf{x}^t} \mathcal{L}(\mathbf{x}^t,\mathbf{y};\:\bm{\theta})}{\|\bigtriangledown_{\mathbf{x}^t} \mathcal{L}(\mathbf{x}^t,\mathbf{y};\:\bm{\theta})\|_2}} - 1) \Big] \big\}.
\end{equation}
Eq. \eqref{mult_ratio_add} and Eq. \eqref{mult_l2_add} indicate that the multiplicative perturbation can be treated as so-called \textit{signal-dependent additive perturbation}, which involve the input data component in the additive perturbation.

Fig. \ref{fig:noise} shows different types of additive adversarial examples and MultAV examples. We can observe that the additive and the multiplicative perturbation distributions are different. Particularly, there is a clear object contour in the MultAV-$\ell_\infty$ perturbation map, showing the signal-dependency of MultAV. The signal-dependent perturbation is more difficult to deal with since they are related to input data (signals). Because of the uniqueness of MultAV, the defenses tailored to resisting additive attacks may be ineffective, posing a new challenge to video recognition systems.

\section{Experiments}

We apply the proposed MultAV on $\ell_\infty$-norm PGD \cite{madry2018towards}, $\ell_2$-norm PGD, ROA \cite{Wu2020Defending}, AF \cite{zajac2019adversarial} and SPA \cite{lo2020defending} attacks (MultAV-$\ell_\infty$, MultAV-$\ell_2$, MultAV-ROA, MultAV-AF and MultAV-SPA), then evaluate them on adversarial training-based state-of-the-art defense approaches. Furthermore, we look into the visualized feature maps under these adversarial attacks.

\subsection{Exmerimental Setup}
\label{setup}

\begin{figure*}[!htbp]
	\begin{center}
		\centering
		\includegraphics[width=1\textwidth]{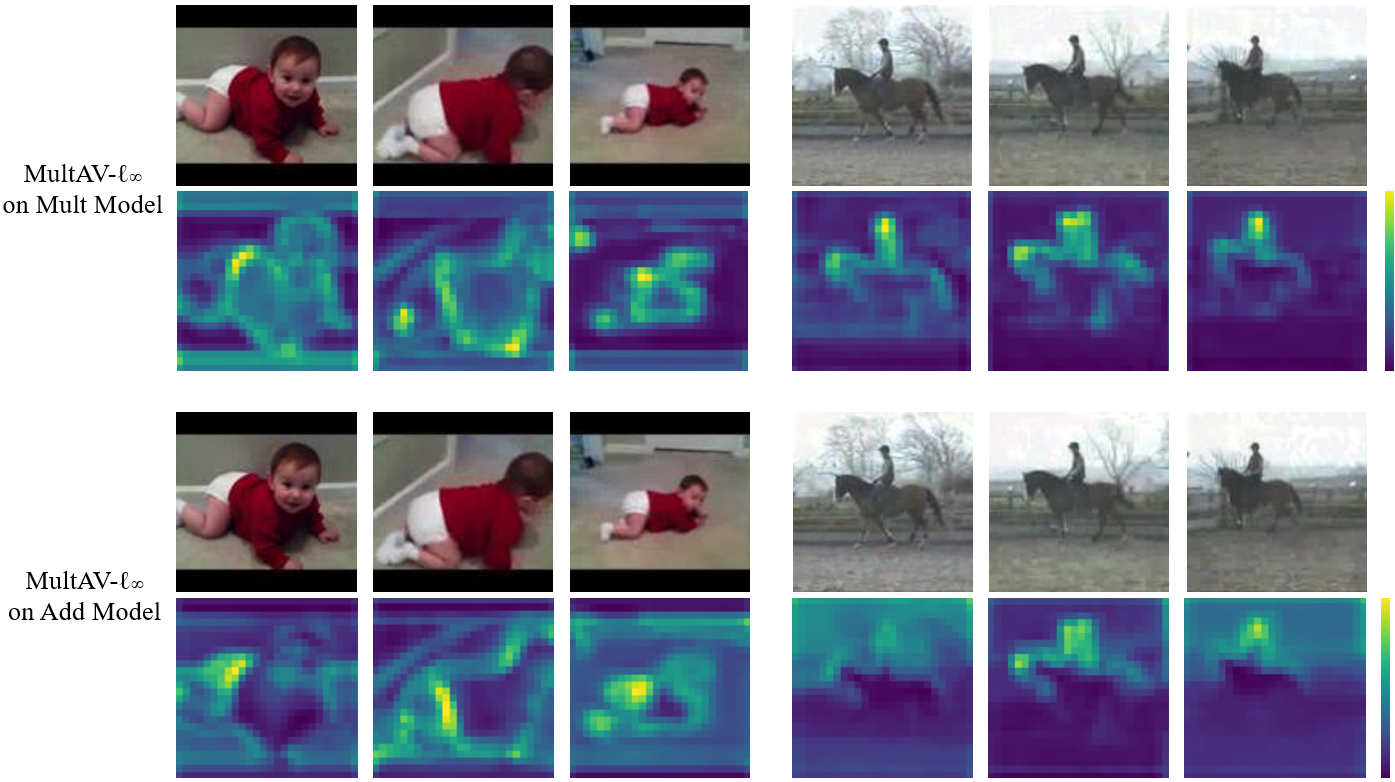}
		\caption{MultAV-$\ell_\infty$ examples generated on Mult Model and Add Model (top); and their corresponding feature maps (bottom). Three frames of the video are displayed here.}
		\label{fig:feature}
	\end{center}
\end{figure*}

We conduct our experiments on the UCF101 dataset\cite{Soomro2012ucf}, an action recognition dataset consisting of 13,320 videos with 101 action classes. We use 3D ResNet-18 \cite{hara3dcnns}, a 3D convolution version of ResNet-18 \cite{he2016deep}, as our video classification network. All the models are trained or adversarially trained by SGD optimizer.

We adversarially train models against different MultAV types respectively and evaluate their robustness to these MultAV examples. The MultAV settings for both inference and adversarial training in our experiments are described as follows:
\begin{itemize}
	\item MultAV-$\ell_\infty$: $\epsilon_m=1.04$, $\alpha_m=1.01$, and $T=5$.
	\item MultAV-$\ell_2$: $\epsilon_m=160$, $\alpha_m=3.55$, and $T=5$.
	\item MultAV-ROA: Rectangle size 30$\times$30, $\epsilon_m=1.7$, $\alpha_m=3.55$, and $T=5$.
	\item MultAV-AF: Framing width 10, $\epsilon_m=3.55$, $\alpha_m=1.7$, and $T=5$.
	\item MultAV-SPA: 100 adversarial pixels on each video frame, $\epsilon_m=3.55$, $\alpha_m=1.7$, and $T=5$.
\end{itemize}

We also adversarially train models against the additive counterparts and evaluate their robustness to MultAV. The additive counterparts for adversarial training are set to similar attacking strength to their corresponding MultAV examples. The settings are as below:
\begin{itemize}
	\item PGD-$\ell_\infty$: $\epsilon=4/255$, $\alpha=1/255$, and $T=5$.
	\item PGD-$\ell_2$: $\epsilon=160$, $\alpha=1.0$, and $T=5$.
	\item ROA: Rectangle size 30$\times$30, $\epsilon=255/255$, $\alpha=70/255$, and $T=5$.
	\item AF: Framing width 10, $\epsilon=255/255$, $\alpha=70/255$, and $T=5$.
	\item SPA: 100 adversarial pixels on each video frame, $\epsilon=255/255$, $\alpha=70/255$, and $T=5$.
\end{itemize}

We test these attack approaches on 3D ResNet-18 with standard training and Madry's adversarial training method \cite{madry2018towards}. Madry's method performs minimax-based adversarial training on an original network architecture. We also evaluate a feature denoising-based state-of-the-art defense approach \cite{xie2019feature}. Feature Denoising adds feature denoising blocks to an original network and performs the same adversarial training protocol as Madry's method on the entire network. We use the Gaussian version of non-local means denoising, which is their top-performing denoising operation. Feature Denoising is designed for only image data, so we extend it to video recognition tasks in two ways: 3D Denoise changes its operations to the 3D domain directly, and 2D Denoise performs the original 2D operations on each video example frame-by-frame. Following the deployment in \cite{xie2019feature}, we insert 3D Denoise/2D Denoise after the conv2, conv3, conv4 and conv5 blocks of 3D ResNet-18.

\subsection{Evaluation Results}

The experimental results are reported in Table \ref{table:results}. Doing adversarial training makes models more robust to MultAV. However, we can see a serious robustness gap between the adversarially trained models against the additive counterparts (Add Model) and against MultAV directly (Mult Model). In particular, Add Model is less robust than Mult Model against MultAV, showing that the defenses tailored to defending against additive treat models fail to fully display their robustness under MultAV. Such gap appears not only across all the considered MultAV types but also across the networks with and without Feature Denoising, which demonstrates the uniqueness and the threat of the proposed MultAV.

The gap size depends on MultAV types. It ranges from 5.39\% to 11.71\% on MultAV-$\ell_\infty$, from 3.81\% to 6.29\% on MultAV-$\ell_2$, and from 0.69\% to 1.67\% on MultAV-ROA. The gap on MultAV-SPA is small. The reason is that the SPA perturbation is composed of scattered single pixels, and such noise distribution has no obvious difference between additive and multiplicative perturbation. Instead, the distribution differences are apparent on the other MultAV types. The most significant gap can be up to 15.12\%, which appears on MultAV-AF.

On the other hand, Feature Denoising \cite{xie2019feature} does not perform well for the video recognition task. For both 3D Denoise and 2D Denoise, their clean data performance and adversarial robustness are degraded as compared with the original architecture. This indicates that an excellent defense for image recognition may be ineffective for videos. Our MultAV can be a good attack method motivating deeper exploration for adversarial robustness in the video domain.

\subsection{Feature Map Visualization}

Fig. \ref{fig:feature} shows the effect of adversarial perturbation on features. As can be seen from this figure, Mult Model is able to capture semantically informative content in video frames. By contrast, Add Model's feature maps are blurred, which means Add Model is easier to be distracted by multiplicative adversarial perturmation and thus cannot focus on semantically informative regions. This visual result is consistent with the quantitive results in Table \ref{table:results}. The proposed MultAV poses a new and strong threat to video recognition models.


\section{Conclusion}

In this paper, we propose a new attack method against video recognition networks, MultAV, which produces multiplicative adversarial videos having different noise distributions to the additive counterparts. It is a general multiplicative algorithm that applies to various attack types ranging from ratio-bounded attacks to physically realizable attacks. It challenges the defense approaches tailored to resisting additive adversarial attacks. Moreover, adversarial robustness in the video domain still lacks exploration. This clearly shows the threat of our MultAV. We hope this work will encourage the research community to look into more general and more powerful defense solutions for video recognition networks.

\section*{Acknowledgment}
This work was supported by the DARPA GARD Program HR001119S0026-GARD-FP-052.

{\small
\bibliographystyle{ieee}
\bibliography{cite}
}

\end{document}